\documentclass[10pt,twocolumn,letterpaper]{article}

\usepackage{cvpr}
\usepackage{times}
\usepackage{epsfig}
\usepackage{graphicx}
\usepackage{amsmath}
\usepackage{amssymb}

\usepackage[breaklinks=true,bookmarks=false]{hyperref}

\cvprfinalcopy 

\def\cvprPaperID{****} 

\begin{document}

\title{M2CAI Workflow Challenge: Convolutional Neural Networks with Time Smoothing and Hidden Markov Model for Video Frames Classification}

\author{Remi Cadene \hspace{1cm} Thomas Robert \hspace{1cm} Nicolas Thome \hspace{1cm} Matthieu Cord\\
Sorbonne Universites, UPMC Univ Paris 06, CNRS, LIP6 UMR 7606, 4 place Jussieu, 75005 Paris\\
{\tt\small \{remi.cadene, thomas.robert, nicolas.thome, matthieu.cord@lip6.fr\}}
}

\maketitle

\begin{abstract}
Our approach is among the three best to tackle the M2CAI Workflow challenge. The latter consists in recognizing the operation phase for each frames of endoscopic videos. In this technical report, we compare several classification models and temporal smoothing methods. Our submitted solution is a fine tuned Residual Network-200 on 80\% of the training set with temporal smoothing using simple temporal averaging of the predictions and a Hidden Markov Model modeling the sequence.
\end{abstract}

\section{Introduction}

The \emph{M2CAI workflow} challenge \cite{twinanda2016endonet, sahu2016, dergachyova2016, jin2016} consists in analyzing endoscopic videos of
minimally invasive surgical operations of cholecystectomy. The task at hand is
to detect at which of the 8 phases of the operation each frames belong. This can
be useful to evaluate a surgeon or to trigger automatic actions in the operating
room for example.

For this challenge, the task must be carried out ``online'', meaning the prediction
for a given frame can only be made based on past frames and predictions, without
any knowledge of the future. This simulates a process of prediction in real-time.

The dataset \cite{Stauder2016} consists of 27 videos in the training set, ranging from 14 to 66
minutes and a test set of 15 videos. The videos have a resolution of 1920 $\times$
1080 pixels and are shot at 25 frames per second at the IRCAD research center in
Strasbourg, France.

In this paper, we describe the approach submitted for the competition that yields a Jaccard index of 71.9. Furthermore, to explain the choices leading to this approach, we compare different models on our validation set. Our experiments can be reproduce using our code : \url{http://github.com/Cadene/torchnet-m2caiworkflow}. Thus, we will not delve into too much details about our implementation.

\subsection{Our approach}

The approach described in this paper consists of two main step, one step of visual recognition without any temporal component using a deep Convolutional Neural Network (CNN), and one step of temporal smoothing to improve the coherence of the predicted sequence using averaging and a Hidden Markov Model (HMM).

\section{Visual recognition: Deep CNN}

The first step of our approach is to learn a classification model of video frames.

To do so, we split randomly the training data into two sets of videos, a full training set of 22 videos and a validation set of 5 videos. In our approach, the training set is made of the following videos: 1,3,4,5,6,7,8,11,12,14,15,16,17,18,19,20,21,22,23,24,25,26. The validation set is made of: 2,9,10,13,27.

Secondly, we extract one frame every 25 frames, returning 1 frame per second of the video. This is done both on the full training set and testing set (the one without any label).

Thirdly, we train our frames classifier and validate it on our validation set. In this study, we compare several approaches detailed in the following subsections.

\paragraph{Note about online data augmentation}
Data augmentation is known to be important to achieve better local invariance and to regularize the model.
In this challenge, we use online data augmentation. It means that after loading a frame during a training epoch or a testing epoch, we rescale the frame to a random size. Then, we randomly crop a part of the frame corresponding to the input size of the model. Finally, we normalize by channel the resulting crop using the corresponding mean and standard-deviation vectors. 

\subsection{Pre-trained CNN as Features Extractor}

This approach consists in using a pre-trained CNN to vectorize the training and validation sets and to train a multi-class classifier on the features vectors, such as a linear model with a cross entropy criterion or Support Vector Machines with a one-versus-all strategy.
Usually the features are extracted at the end of the CNN after a fully connected layer or an activation functions. This is the usual baseline method for transfer learning \cite{simonyan2014very}.

In this challenge, we use features extracted from the penultimate layer of InceptionV3 \cite{szegedy2015rethinking} (2048 dimensions) and train a linear model with a cross entropy criterion. Our preprocessing is as follow. Given that InceptionV3 takes as input images of size 299x299, we randomly rescale the original images between 299 and 330 pixels of width or height based on the smallest fitting dimension. Then, we randomly crop a square region of size 299x299. Finally, we normalize each pixels using the mean and standard deviation processed on ImageNet.


\subsection{Fine Tuning Pre-trained CNN}

Fine tuning consists in training a pre-trained CNN on a smaller dataset. Typically, the last fully connected layers, which can be viewed as classification layers, are reset and a smaller learning rate is applied to the pre-trained layers. By doing so, the goal is to adapt the representations learned to the new dataset. The more different is the latter from the original dataset, the more layers must be reset.

In this study, we remove the last layer of a pre-trained InceptionV3 on ImageNet and add a new fully connected of output size equal to 8, the number of classes in M2CAI Workflow. We use Adam \cite{Kingma14} as optimizer. It is robust regarding the learning rate factor for pre-trained layers. Thus, we fix the latter to 10 times smaller. We use grid search to find the best set of hyperparameters : the learning rate and the learning rate decay (which decreases the adaptive learning rate of each parameters by a factor after each forward-backward pass).

We also fine-tune an other pre-trained CNN on ImageNet called Residual Network-200 \cite{he2016deep}. It takes as input images of size 221x221. Thus, we use the same preprocessing with a minimal size and a crop size of 221 and a maximum size of 240.

\subsection{Other Baselines}

The first baseline consists in training InceptionV3 from scratch. It means that we reset all its parameters with LSUV method \cite{LSUVInit2015}. We use the same preprocessing than before. However, we process the mean and std over our training set.

The second baseline is to fine-tune an Inception-V3 with the WELDON aggregation layer plugged at the end \cite{durand2016weldon}.
It is known to be effective in the case of small datasets with strong spatial variance, specifically objects translationally invariant in the image.
We use the same preprocessing than for fine tuning. However, we use a minimal image size and crop size of 448 and a maximum size of 463. Hence, the input of the spatial aggregation layer is a spatial map of features and no longer a vector of features.

\section{Temporal smoothing: Averaging and HMM}

\subsection{Averaging}

Using a model defined in the previous section, we obtain for each images a vector of log-probabilities.
To temporally smooth the predictions, we average the vectors across the
last 15 frames (corresponding to 15 seconds of the video). This means that our
smoothed prediction has a lag of 7.5 seconds. However, the metric of the challenge
allows a 10-second-margin in the predictions, meaning that it is not considered
problematic to have a slight lag in our prediction. We therefore take advantage
of this tolerance to improve the smoothness of the predictions.

\subsection{Hidden Markov Model}

In addition to the averaging, we propose to use an HMM to model the transitions
between the various steps of the operation. To do so, we consider that the step of the operation is the discrete hidden state $x_t$,
from which we only observe a noisy vector $y_t$ that is our averaged vector of log-probabilities from
the network.

\paragraph{Training} An HMM has 3 kind of parameters: the initial state probabilities, the matrix of probabilities of transition between states from one time-step to the next, and the parameters regarding emission of observations for each possible step.

We compute those information on the training set. The initial state probabilities and the matrix of transition are computed by simple counting. We chose to model the emission of
observation with a Gaussian distribution. To do so, we compute for each step the average observation and its covariance matrix.

\paragraph{Offline prediction} In this study, the transition matrix of our trained HMM is sparse, which imposes a lot of structure on the predicted sequence of states. By applying Viterbi algorithm, we ``decode'' a sequence of observations to obtain the most likely sequence of states. This is done for our offline prediction.

\paragraph{Online prediction} However, for this challenge, the predictions must
be given in online mode. Therefore, to predict the state $x_t$, we apply the Viterbi algorithm on the sequence $y_1,...,y_t$ and keep the last state of the predicted sequence. This process ensure that we are working in online mode. However, it decreases the performance of the HMM, because the constrains on the transition matrix are no longer enforced on the final output sequence.

\subsection{Post-processing to produce requested files}

As for now, our two-step model produces predictions at a rate of 1 frame per second. To come back to the required rate of 25 fps, we simply copy each predictions 25 times, and
then crop or pad the predicted sequence to match exactly the number of frames in the video.

\section{Experiments}

\subsection{Classification models}

In table 1, we compare our classification models by accuracy (top 1) on the validation set. We can see that Extraction is less accurate than From Scratch, which performs worth than Fine Tuning. It means that using the representations computed on ImageNet is possible, even if this dataset is semantically far from M2CAI Workflow. Nevertheless, it is less accurate than learning all the representations from a CNN randomly initialized. However, combining both approaches by adapting the representations learned on ImageNet for this task with fine tuning is the method of choice. Also, Weldon, a better model in case of strong translation invariance, achieves slightly worst results than the classical Fine Tuning. As we do not have any prior knowledge for this task, testing this kind of model is an easy way to question the capacity of classical CNNs to learn translation invariance on this dataset. Finally, we take our best model, namely ResNet200 Fine-tuned, as our frames classifier.

\begin{table}
	\begin{center}
		\begin{tabular}{|c|c|}
			\hline
			Classification Model & Accuracy (\%) \\
			\hline\hline
			InceptionV3 Extraction & 60.53 \\
			InceptionV3 From Scratch & 69.13 \\
			InceptionV3 Weldon & 78.18 \\
			InceptionV3 Fine-tuned & 79.06 \\
			ResNet200 Fine-tuned & 79.24 \\
			\hline
		\end{tabular}
	\end{center}
	\caption{Accuracy on the validation set.}
\end{table}

\subsection{Temporal methods}

In table 2 and 3, we compare our temporal methods on the validation set. As expected, an HMM trained on top of the smoothed predictions achieves better results. However, we keep the online predictions of our HMM for the final submission produced on the testing set.

\begin{table}
	\begin{center}
		\begin{tabular}{|c|c|c|}
			\hline
			Temporal Method & Accuracy (\%) & Jaccard \\
			\hline\hline
			Avg Smoothing & 85.97 $\pm$ 3.75 & 74.67 $\pm$ 7.87\\
			HMM Online & 88.90 $\pm$ 3.55 & 81.60 $\pm$ 10.49\\
			HMM Offline & 93.47 $\pm$ 3.59 & 87.59 $\pm$ 6.97\\
			\hline
		\end{tabular}
	\end{center}
	\caption{Accuracy Top1 and Jaccard score on the validation set. The variance is computed over all classes.}
\end{table}

\begin{table*}
	\begin{center}
		\begin{tabular}{|c|c|c|c|}
			\hline
			Temporal Model by classes & Avg Smoothing & HMM Online & HMM Offline \\
			\hline \hline
			0.TrocarPlacement & 74.92 & 99.19 & 99.82 \\
			1.Preparation & 73.51 & 83.69 & 85.75 \\
			2.CalotTriangleDissection & 84.15 & 85.85 &  94.19 \\
			3.ClippingCutting & 76.25 & 74.68 & 82.51 \\
			4.GallbladderDissection & 69.82 & 69.21 & 81.38 \\
			5.GallbladderPackaging & 76.06 & 67.71 & 79.89 \\
			6.CleaningCoagulation & 83.41 & 86.88 & 91.62 \\
			7.GallbladderRetraction & 59.25 & 85.61 & 85.53 \\
			\hline
			All classes & 74.67 & 81.60 & 87.59 \\
			\hline
		\end{tabular}
	\end{center}
	\caption{Jaccard score on the validation set.}
\end{table*}

In figure 1, we compare our three temporal methods by accuracy error (100 - accuracy top 1). We plot the predicted classes over the frames of each videos. We can see that using an HMM is really effective to add a second temporal smoothing.

\begin{figure*}
\begin{center}
   \includegraphics[width=1\linewidth]{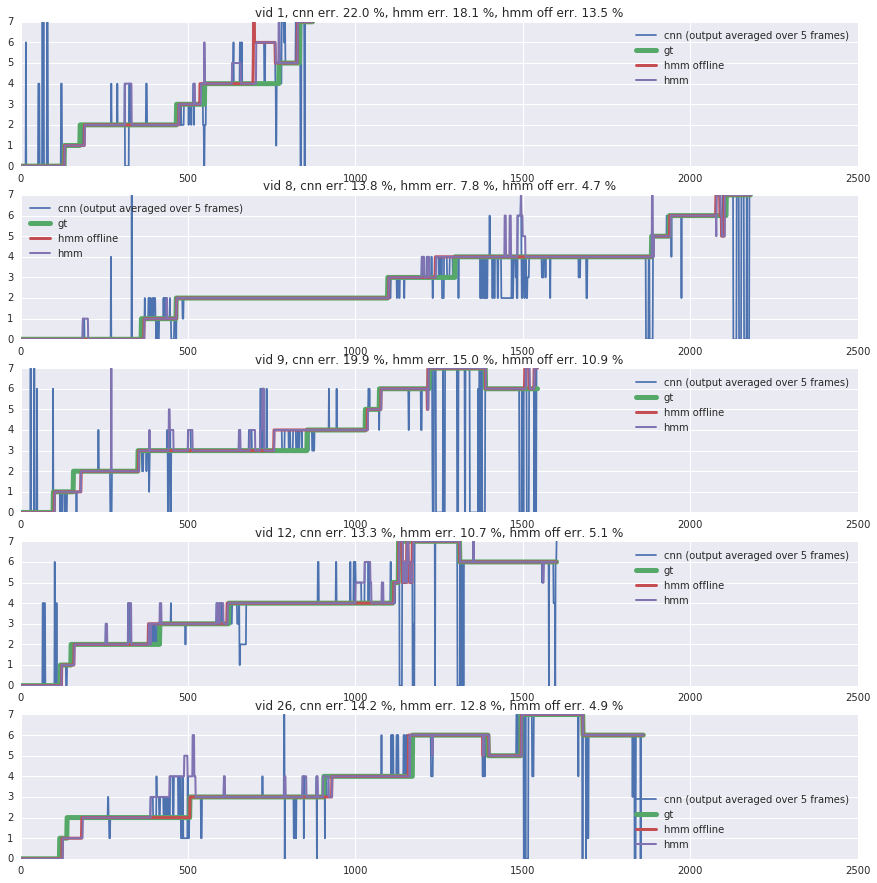}
\end{center}
   \caption{Comparison of our temporal models predictions on the validation set. In blue, our ResNet200 Fine-tuned with average smoothing over 5 frames. In red, the offline predictions of an HMM trained of top of the latter model predictions. In mauve, the online predictions. In green, the ground truth label. }
\label{fig:long}
\label{fig:onecol}
\end{figure*}

\section{Conclusion}

In this challenge, we tried several classification models and temporal smoothing methods to classify in an online meaner frames of surgical videos. We validated that fine tuning Convolutional Neural Networks is a good approach on this dataset. In order to improve our classification results, we wanted to try two approaches that we did not have the time to submit before the end of the challenge. The first method consists in fine tuning the same CNNs with the same hyperparameters (learning rate, early stopped epoch, etc.) on the full training set. The second method consists in fine tuning multiple CNNs on different training videos and then of using an ensembling approach such as a vote by majority to produce more stable predictions. Finally, we did not have the time to look at the frames of video poorly classified. This last step could helped us to understand the difficulties experienced by our model in order to improve it.

{\small
\bibliographystyle{ieee}
\bibliography{egbib}
}

\end{document}